\documentclass[conference]{IEEEtran}

\usepackage{amsmath}
\usepackage{amssymb}
\usepackage{color}
\usepackage{hyperref}
\usepackage{lettrine}

\newcommand*\hexbrace[2]{%
  \underset{#2}{\underbrace{\rule{#1}{0pt}}}}

\begin{document}

\title{Optimal V2G Scheduling of Electric Vehicles and Unit Commitment using Chemical Reaction Optimization}

\author{James J.Q. Yu,
       \textit{Student Member, IEEE}\\
       Department of Electrical and\\
       Electronic Engineering\\
       The University of Hong Kong\\
       Email: jqyu@eee.hku.hk\\
\and Victor O.K. Li,
       \textit{Fellow, IEEE}\\
       Department of Electrical and\\
       Electronic Engineering\\
       The University of Hong Kong\\
       Email: vli@eee.hku.hk\\
\and Albert Y.S. Lam,
       \textit{Member, IEEE}\\
       Department of Computer Science\\
       Hong Kong Baptist University\\
       Email: albertlam@ieee.org\\
}

\maketitle
\pagestyle{empty}

\begin{abstract}
An electric vehicle (EV) may be used as energy storage which allows the bi-directional electricity flow between the vehicle's battery and the electric power grid. In order to flatten the load profile of the electricity system, EV scheduling has become a hot research topic in recent years. In this paper, we propose a new formulation of the joint scheduling of EV and Unit Commitment (UC), called EVUC. Our formulation considers the characteristics of EVs while optimizing the system total running cost. We employ Chemical Reaction Optimization (CRO), a general-purpose optimization algorithm to solve this problem and the simulation results on a widely used set of instances indicate that CRO can effectively optimize this problem.
\end{abstract}


\section{Introduction}

\lettrine[lines=2]{W}{ith} the growing concern on global climate change, governments and industries have invested extensively in environmentally friendly  technologies. The transportation sector is responsible for a large portion (24\%) of green house gas emission \cite{book/wiley/SLabatt2007}, which has been recognized as one of the major cause of global climate change. To alleviate such emissions, incentives have been provided to encourage the adoption of electric vehicles (EVs). The next-generation EVs have drawn the interest of researchers in recent years, as they have the capability of performing vehicle-to-grid (V2G) operation \cite{journals/tsg/YifengHe2012}. V2G technology  \cite{journals/ep/HLund2008} is regarded as an important application of smart grid technology. An EV may be used as energy storage which allows the bi-directional electricity flow between the vehicle's battery and the electric power grid \cite{LamLeungLiCapacityManagementVehicle}\cite{LamHuangSilvaSaad2012MultilayerMarket}. V2G can efficiently flatten the load profile of the electric system with optimal scheduling of charging (grid-to-vehicle, G2V) and discharging (V2G) behavior, which can potentially reduce the total system running cost and green house gas emission \cite{journals/tcst/DongsukKum2013}.

Recently, a number of algorithms for scheduling the charging and discharging of electric vehicles have been proposed \cite{journals/tsg/YifengHe2012}\cite{conference/noms/KMets2010}\cite{conference/e2030/CHutson2008}\cite{journals/tie/AYSaber2011}\cite{journals/tsj/AYSaber2012}.
However, the algorithms proposed in \cite{conference/noms/KMets2010} and \cite{conference/e2030/CHutson2008} only consider EV charging during the scheduling process. Though \cite{journals/tie/AYSaber2011} and \cite{journals/tsj/AYSaber2012} involve V2G operation to minimize the cost, the consideration of system constraints, especially the EV-related constraints, is inadequate. In particular, they fail to take the features of EVs into consideration. The algorithm proposed in \cite{journals/tsg/YifengHe2012} is efficient in reducing the EV individual cost, but the authors did not take the system running cost into account. In particular, the Unit Commitment (UC) problem, or the scheduling of generator units, is ignored. In order to avoid these drawbacks and provide an integrated solution of the complete power system, we introduce UC into the optimal scheduling model and propose a new formulation of jointly scheduling of Electric Vehicle and Unit Commitment (EVUC).

Metaheuristic is a kind of general-purpose algorithm which optimizes problems in an iterative manner, trying to find or improve a candidate solution given a measure of quality. It is a very popular approach to solve UC-related problems \cite{journals/tie/AYSaber2011}\cite{journals/tps/PHChen2012}. Among all metaheuristics, Chemical Reaction Optimization (CRO) is a promising algorithm in solving combinatorial and continuous optimization problems \cite{journals/tec/AYSLam2010}. CRO mimics the behaviour of molecules in a chemical reaction. It has been used effectively in solving many real-world problems \cite{conference/cec/JYu2011ANN}\cite{conference/cec/JYu2012Sensor}. In this work, we use CRO to find optimal solutions of our proposed EVUC problem.

The main contribution of this paper is a new formulation of the joint scheduling of V2G and UC. Compared with previous formulations, our new formulation introduces additional constraints to make it more practical. We also perform simulations to demonstrate that CRO is a good method for solving this problem.

The rest of the paper is organized as follows. The related work is presented in Section II. Section III introduces the nomenclature we use in this paper. Section IV formulates the EVUC problem and the implementation of CRO to solve this problem is described in Section V. We will demonstrate the experiment instance and the simulation results in Section VI, accompanied with analysis and discussion. Finally we will conclude this paper in Section VII.

\section{Related Work}

Existing work of V2G operation scheduling of EVs can be divided into two classes: charging-only scheduling and bi-directional scheduling. In charging-only scheduling, the algorithms try to optimize the electricity flow from the power grid to the batteries of EVs. For example, Shrestha \textit{et al.} optimized the EV charging cycles to off-peak periods to flatten the demand curve, in order to reduce the charging cost \cite{conference/ipec/Shrestha2007}. Mets \textit{et al.} presented a smart energy control strategy to charge residential plug-in hybrid EVs (PHEVs) to smooth the system load profile  \cite{conference/noms/KMets2010}. However, with the development of V2G technology, bi-directional charging, i.e., V2G and G2V, is possible, and bi-directional scheduling algorithm has attracted much research recently. The role of EVs in the power system may change during the day from loads to sources, and vice versa. Binary particle swarm optimization was employed to tackle the V2G scheduling problem to minimize the total running cost and reduce green house gas emission in \cite{journals/tie/AYSaber2011} and \cite{journals/tsj/AYSaber2012}. Han \textit{et al.} proposed an aggregator for V2G frequency regulation in \cite{journals/tsg/SHan2010}, aiming to maximize the revenue.

CRO is a recently proposed metaheuristic, which has been developed intensely in the past few years. CRO was originally designed to solve combinatorial optimization problems in \cite{journals/tec/AYSLam2010}, where CRO is adopted to solve the Quadratic Assignment Problem, the Resource-Constraint Project Scheduling Problem, and the Channel Assignment Problem. The Cognitive Radio Spectrum Allocation Problem is addressed in \cite{conference/globecom/AYSLam2010}. Yu \textit{et al.} proposed and solved a Sensor Deployment Problem with CRO in \cite{conference/cec/JYu2012Sensor}. Lam \textit{et al.} analyzed the convergence of CRO for combinatorial optimization in \cite{journals/tec/AYSLam2013Early}. Lam \textit{et al.} also proposed Real-Coded CRO, a variant of CRO, to solve continuous optimization problems in \cite{journals/tec/AYSLam2012}. Yu \textit{et al.} solved an Artificial Neural Network training problem in \cite{conference/cec/JYu2011ANN}, and proposed several perturbation functions for RCCRO in \cite{conference/cec/JJQYu2012PF}.

Researchers have been using metaheuristics to solve UC and its related problems for many years. Mantawy \textit{et al.} proposed a hybrid algorithm integrating genetic algorithm, tabu search and simulated annealing to solve UC in \cite{journals/tps/Mantawy1999}. Rajan \textit{et al.} proposed an evolutionary programming-based tabu search method for the same problem in \cite{journals/tps/Rajan2004}. Yousuf \textit{et al.} proposed a binary particle swarm optimization to solve the UC with renewable energy sources in \cite{journals/tsj/AYSaber2012}. Chen proposed an expert system with elite particle swarm optimization algorithm to solve UC in \cite{journals/tps/PHChen2012}. As CRO has been applied to solve related power system optimization problems, e.g.,  \cite{SunLamLiXuYu2012ChemicalReactionOptimization}\cite{XuWenLiLeung2013OptimalPMUPlacement} and has demonstrated outstanding performance, we adopt CRO to solve this EVUC problem.

\section{Nomenclature}
\addcontentsline{toc}{section}{Nomenclature}
\begin{IEEEdescription}[\IEEEusemathlabelsep\IEEEsetlabelwidth{$a_i,b_i,c_i$}]
\item[$T$] Total number of time intervals.
\item[$t$] The index of a time interval.
\item[$\Delta t$] Length of a time interval.
\item[$I$] Total number of thermal units.
\item[$i$] The index of a thermal unit.
\item[$M$] Total number of EVs.
\item[$m$] The index of an EV.
\item[$P^t_i$] Power output of unit $i$ at time $t$.
\item[$U^t_i$] State of unit $i$ at time $t$. 1 is online and 0 is offline.
\item[$f^{FC}_i(P)$] Fuel cost of unit $i$ when generating $P$ power output.
\item[$UC_i$] Start-up cost of unit $i$.
\item[$DC_i$] Shut-down cost of unit $i$.
\item[$a_i,b_i,c_i$] Fuel cost coefficients of unit $i$.
\item[$\overline{P_i}$] Maximum power output of unit $i$.
\item[$\underline{P_i}$] Minimum power output of unit $i$.
\item[$\mathcal{T}_{i,mr}$] The set of time intervals when unit $i$ must be online.
\item[$\mathcal{T}_{i,mo}$] The set of time intervals when unit $i$ must be offline.
\item[$P^t_D$] System load demand at time $t$.
\item[$P^t_{SR}$] Spinning reserve at time $t$.
\item[$P^t_{EV}$] The amount of power discharged from EV through V2G at time $t$. A positive value represents discharging to the power grid (V2G) and a negative value represents charging from the power grid (G2V).
\item[$MUT_i$] Minimal uptime of unit $i$.
\item[$MDT_i$] Minimal downtime of unit $i$.
\item[$\tau^t_i$] The number of continuous online or offline time intervals before time $t$ for unit $i$. A possitive value represents online state and a negative represents offline state.
\item[$URR_i$] Maximum up-ramp rate limit of unit $i$.
\item[$DRR_i$] Maximum down-ramp rate limit of unit $i$.
\item[$E^{\textit{cap}}_m$] Battery capacity of EV $m$.
\item[$E^t_m$] The amount of electricity hold by EV $m$ at time $t$.
\item[$\textit{freq}_m$] Charging frequency of EV $m$.
\item[$\mathcal{T}^{\textit{charge}}$] The set of time intervals when EVs are charging from the power grid $\mathcal{T}^{\textit{charge}}=\{t|\forall P^t_{EV}<0\}$.
\item[$E^{\textit{con}}_m$] Total electricity consumed by EV $m$ in a complete scheduling period.

\end{IEEEdescription}

\section{EVUC Problem Formulation}\label{sec:problem}

The purpose of UC problem is to determine the schedule of the start up and shut down of power generator units, such that the total power output meets the fluctuating load over the scheduling period at minimal cost \cite{journals/tps/PHChen2012}. EVs connected to the grid can act as loads, sources, or energy storages. The EVUC problem can be formulated as a constrained nonlinear optimization problem if we divide the scheduling period into time intervals as follows:
\begin{equation}\label{eqn:objective}
\begin{aligned}
\min_{P^t_i}\sum^T_{t=1}\sum^I_{i=1}[f^{FC}_i(P^t_i)U^t_i&+UC_i(1-U^{t-1}_i)U^t_i\\
&+DC_iU^{t-1}_i(1-U^t_i)],\\
\text{over }P^t_i\text{ for }i=1,2,\cdots&,I,\hspace{0.5em}t=1,2,\cdots,T.
\end{aligned}
\end{equation}

In power systems, the fuel cost of a thermal unit is usually formulated as a quadratic function:
\begin{equation}\label{eqn:fuelcost}
f^{FC}_i(P)=a_i+b_iP+c_iP^2.
\end{equation}

This objective function of EVUC is subject to two classes of constraints: UC and EV constraints. The former constraints are introduced by the original UC problem \cite{journals/tps/PHChen2012} and the latter ones are introduced due to the special characteristics of EVs.

\subsection{UC Constraints}

When considering UC constraints, we can consider the collection of EVs as a new type of unit which can generate or consume power at different times.

\subsubsection{Generation Constraints}

Every online unit has generation limits:
\begin{equation}\label{eqn:generationconst}
\underline{P_i}\leq P^t_i\leq \overline{P_i}\hspace{1em}i=1,2,\cdots,I,\hspace{0.5em}t=1,2,\cdots,T.
\end{equation}

\subsubsection{Must-run and Must-off Units}

Sometimes units are assigned to be in a must-run or must-off status to meet different requirements:
\begin{equation}\label{eqn:mrandmo}
\begin{aligned}
U^t_i=1\hspace{1em}\text{for } t\in\mathcal{T}_{i,mr}\\
U^t_i=0\hspace{1em}\text{for } t\in\mathcal{T}_{i,mo}
\end{aligned}.
\end{equation}

\subsubsection{System Power Balance}

The generation and demand of the system must be identical:
\begin{equation}\label{eqn:powerbalance}
\sum^I_{i=1}P^t_iU^t_i+P^t_{EV}-P^t_D=0\hspace{1em}t=1,2,\cdots,T.
\end{equation}

\subsubsection{Spinning Reserve Constraints}

In order to prevent power supply interruptions, an adequate amount of spinning reserve is essential for a power system:
\begin{equation}\label{eqn:spinningreserve}
\begin{aligned}
\sum^I_{i=1}\overline{P_i}U^t_i+P^t_{EV}-P^t_D-&P^t_{SR}\geq0\\
&\hspace{1em}t=1,2,\cdots,T
\end{aligned}.
\end{equation}

\subsubsection{Minimal Uptime and Downtime}

A unit must be online or offline for a certain number of time intervals before it can be shut down or started up:
\begin{equation}\label{eqn:mutmdt}
\begin{aligned}
\tau^t_i\geq\textit{MUT}_i\times &U^{t-1}_i(1-U^t_i)\\
-\tau^t_i\geq\textit{MDT}_i\times &(1-U^{t-1}_i)U^t_i\\
&\hspace{1em}i=1,2,\cdots,I,\hspace{0.5em}t=1,2,\cdots,T
\end{aligned}.
\end{equation}

\subsubsection{Ramp Rate Limit}

A unit cannot change its power output too rapidly. The range is constrained by the ramp rate limits:
\begin{equation}\label{eqn:ramprate}
\begin{aligned}
P^t_i-P^{t-1}_i\leq \textit{URR}_i\hspace{1em}i=1,2,\cdots,I,\hspace{0.5em}t=1,2,\cdots,T\\
P^{t-1}_i-P^t_i\leq \textit{DRR}_i\hspace{1em}i=1,2,\cdots,I,\hspace{0.5em}t=1,2,\cdots,T
\end{aligned}.
\end{equation}

\subsection{EV Constraints}

\subsubsection{Capacity Limit}

The total amount of electricity which can be stored in the EVs is limited by the capacity of the batteries in the EVs:
\begin{equation}\label{eqn:capacity}
\sum^M_{m=1}E^{\textit{cap}}_m-\sum^M_{m=1}E^t_m\geq0\hspace{1em}t=1,2,\cdots,T.
\end{equation}

\subsubsection{Charging Frequency Limit}

In order to save the battery life, it is suggested to limit the charging frequency of EVs \cite{journals/tie/Chen2013}. So the maximum amount of electricity charged to EVs is limited:
\begin{equation}\label{eqn:chargefreq}
\sum_{t\in\mathcal{T}^{\textit{charge}}}P^t_{EV}\times\Delta t\leq\sum^M_{m=1}(E^{\textit{cap}}_m\times\textit{freq}_m).
\end{equation}

\subsubsection{Battery Electricity Balance}

The total electricity stored in the batteries of EVs shall remain the same after a complete scheduling period, otherwise the EV system may have all its electricity depleted, or charged to capacity, rendering it incapable of providing regulation service. In this process, the energy consumed by EVs themselves shall also be considered. Assume that the total number of EVs in the system, i.e., $m$, is constant during the scheduling period. This constraint is formulated as follows:
\begin{equation}\label{eqn:batterybalance}
\sum^T_{t=1}P^t_{EV}\times\Delta t+\sum^M_{m=1}E^{\textit{con}}_m=0.
\end{equation}

\section{Algorithm Design}

In this section, we will first briefly review CRO. Then the detailed implementation of our proposed methodology will be presented.

\subsection{A Brief Review of CRO}

CRO mimics the behavior of molecules in a chemical reaction. Consider a closed container with some molecules. Each molecule has a molecular structure, which is used to represent a feasible solution, and different kinds of energy, which represent some solution quality-related parameters. As time evolves, the molecules move around randomly and collide with the container wall or with each other. The collisions modify the molecular structures of participated molecules according to some predefined rules. If the modification caused by the collision accords with the energy conservation law, then the modification is accepted and the molecular structure, i.e., a feasible solution, is potentially improved. CRO utilizes this kind of modifications to perform optimization tasks.

In CRO, there are four kinds of elementary reactions, namely, on-wall ineffective collision (\textit{on-wall}), decomposition (\textit{dec}), inter-molecular ineffective collision (\textit{inter}), and synthesis (\textit{syn}). In each iteration of CRO, only one out of these four elementary reactions will occur. Among these elementary reactions, \textit{on-wall} and \textit{dec} take one molecule as input (parent molecule) while \textit{inter} and \textit{syn} take two molecules. \textit{on-wall} and \textit{syn} employ the input molecule(s) to generate one output (child molecule) while \textit{dec} and \textit{inter} generate two. The occurrence of these elementary reactions are controlled by different parameters. Although they are quite different in terms of inputs and outputs, they share a common characteristic which distinguishes CRO with other metaheuristics. All elementary reactions satisfy the energy conservation law, i.e., the energy in the whole system remains the same before and after the elementary reaction. Interested readers can refer to \cite{journals/tec/AYSLam2010}\cite{journals/mc/AYSLam2012} for details.

\subsection{Encoding Scheme}

As stated in Section \ref{sec:problem}, we use $T$ to represent the total number of time intervals and $I$ to represent the set of units. So we can use a $T\times I$ binary matrix to represent the schedule of online status of thermal units, where 1's stand for online and 0's for offline. Besides this typical encoding scheme for a canonical UC problem, we also append a vector of length $T$ to represent the power output of all EVs. So a typical solution $s$ for the EVUC problem is composed of two parts: an UC part and an EV part as follows:
\begin{center}
\[
\begin{array}{c@{}c}
s =
\left[
\begin{array}{ccccc}
U^1_1 & U^1_2 & \cdots & U^1_I & P^1_{EV}\\
U^2_1 & U^2_2 & \cdots & U^2_I & P^2_{EV}\\
\vdots  & \vdots  & \ddots & \vdots & \vdots  \\
U^T_1 & U^T_2 & \cdots & U^T_I & P^T_{EV}\\
\end{array}
\right].
&\\
\hspace{0.5cm}\hexbrace{3.4cm}{\text{UC}}\hexbrace{1.1cm}{\text{EV}}
\end{array}
\]
\end{center}

\subsection{Initial Solution Generation}

As a feasible solution of EVUC can be divided into two parts, we initialize them separately. Instead of randomly generating binary numbers for the UC part (as is usually the case when using metaheuristic to solve other optimization problems), we use a heuristic proposed in \cite{journals/tps/PHChen2012} to generate this part. Note that this heuristic, or so-called ``Expert System Pre-dispatch" cannot guarantee the solutions generated have good performance. For the EV part, we will dispatch the EV charges evenly without violating the constraints.

The main idea of the initial solution generator of the UC part proposed in \cite{journals/tps/PHChen2012} is that an initial solution will go through all UC constraints to check whether any violation occurs. When a solution violates any constraint, it will be repaired using some predefined ``rules". This process can be further divided into three steps: a) check Constraint (\ref{eqn:mrandmo}), b) check Constraint (\ref{eqn:spinningreserve}), and c) check Constraint (\ref{eqn:mutmdt}). Other UC constraints will be satisfied in the process of Economic Dispatch (ED), which will be introduced later. Interested reader can refer to \cite{journals/tps/PHChen2012} for details of this initial solution generation heuristic.

However, this method has a serious drawback. As the steps previously stated are performed sequentially, it is highly likely that the repair function in Step c may potentially make the solution violate Constraint (\ref{eqn:spinningreserve}) again, despite this solution has just passed the checks in Steps a and b. Here is an example. Suppose a thermal unit $i$ with $MUT_i=MDT_i=3$. The unit state of unit $i$ in a solution which just passed Step a and b check is $[\cdots,1,1,1,\underline{1},0,0,\underline{1},1,1,1,\cdots]^\intercal$. As this sequence does not satisfy $MDT_i$, either state of the two underscored time intervals must be changed to 0. However, this change potentially decreases the maximum power output of this time interval, which may in return violate Constraint (\ref{eqn:spinningreserve}). Moreover, as there is no feedback scheme in this method, this violation will still be retained without repair and the solution becomes infeasible. In order to overcome this drawback, we add a simple recursive scheme to the original method: every solution after going through all the steps will go through these checks again sequentially until no violation is found. As this recursion is a time-consuming task, we will discard this solution and generate a new one if the solution still cannot pass all the constraint checks after 10 recursions. Thus we can guarantee the UC part of our generated initial solution satisfy Constraints (\ref{eqn:mrandmo}), (\ref{eqn:spinningreserve}), and (\ref{eqn:mutmdt}).

For the EV part of an initial solution, we suppose that no smart operation, i.e., having EVs as storages or sources, occurs in the scheduling period. So the total amount of electricity charged to the EVs is $\sum^M_{m=1}E^{\textit{con}}_m$. This amount is first evenly distributed to all time intervals. Then we check this solution against Constraint (\ref{eqn:spinningreserve}), which is the only constraint this solution may violate. If this constraint is violated, then we calculate the excessive electricity $E_{ex}$ at time $t_{ex}$ when the maximum power output of all online thermal units cannot satisfy the requirement of demand and spinning reserve:
\begin{equation}\label{eqn:excessev}
\begin{aligned}
E_{ex}=P^t_D+&P^t_{SR}-P^t_{EV}-\sum^{|\mathcal{I}|}_{i=1}\overline{P_i}U^t_i.
\end{aligned}
\end{equation}
This excessive electricity then is divided evenly and dispatched to all time intervals whose maximum power outputs can satisfy the spinning reserve requirement. This process repeats until no time interval violates constraint (\ref{eqn:spinningreserve}). Here is an example. Suppose there are three thermal units with $\overline{P}=[100,100,100]^\intercal$, the demands and spinning reserves of the three time intervals are $P^t_D+P^t_{SR}=[80,290,170]^\intercal$, and $\sum^M_{m=1}E^{\textit{con}}_m=50$. After the first step of even dispatch, a possible solution is:
\begin{center}
\[
\left[
\begin{array}{cccc}
1&0&0&-16.667\\
1&1&1&-16.667\\
1&1&0&-16.667
\end{array}
\right]
\Rightarrow
\left[
\begin{array}{cccc}
1&0&0&-20\\
1&1&1&-10\\
1&1&0&-20
\end{array}
\right].
\]
\end{center}
However, the excessive electricity in the second time interval violates Constraint (\ref{eqn:spinningreserve}). So the excessive electricity $E_{ex}=290-(-16.667)-300=6.667$ is dispatched to the other two time intervals, rendering the solution feasible. Up to now there is no smart operation in our initial solution. The Constraints (\ref{eqn:capacity}), (\ref{eqn:chargefreq}), and (\ref{eqn:batterybalance}) are naturally satisfied, otherwise the EVs in the system would not have enough electricity to function, and this is not an acceptable situation.

\subsection{Neighborhood Search Operator}

The neighborhood search operator, which modifies one feasible solution and attempts to find another one, is employed in all four elementary reactions in our CRO implementation for this problem. As each solution can be divided into two parts, we will modify them separately.

\subsubsection{UC Part Modification}

At the beginning, the neighborhood search operator will first generate a random position in the $T\times I$ binary matrix except those must-run and most-off positions. The state in this position is then toggled, i.e., $U^t_i\leftarrow1-U^t_i$. Then the newly generated solution will be checked against Constraints (\ref{eqn:spinningreserve}) and (\ref{eqn:mutmdt}). If either one of the constraints is violated, the modification is discarded and the solution is reverted to the original state. In such cases, the algorithm will go on to modify the EV output values. However, if the algorithm successfully modifies one position in the UC part without violating the constraints, the EV output values will not be changed. An example of the operation of this neighborhood search operator is as follows:
\begin{center}
\[
\left[
\begin{array}{cccc}
1&\textbf{0}&0&-20\\
1&1&1&-10\\
1&1&0&-20
\end{array}
\right]
\Rightarrow
\left[
\begin{array}{cccc}
1&\textbf{1}&0&-20\\
1&1&1&-10\\
1&1&0&-20
\end{array}
\right]
\]
\end{center}
where the neighborhood search operator toggles the state of the second unit on the first time interval from offline to online, which is bolded in the above transformation.

\subsubsection{EV Part Modification}

If the UC part modification does not successfully change any state, the algorithm will modify the EV part values. In order not to violate Constraint (\ref{eqn:batterybalance}), the sum of all EV output values shall keep unchanged. So we first select two random time intervals $t_{inc}$ and $t_{dec}$ from $T$, assign one of them to be the time interval for which we decide to increase the EV output (increase V2G or decrease G2V), and the other to be the time interval for decreasing the EV output. As Constraint (\ref{eqn:spinningreserve}) limits the maximum power that the selected outputs can increase/decrease, we first determine the increase/decrease range $r$ as
\begin{equation}\label{eqn:incrange}
\begin{aligned}
r=&\min(P^{t_{\textit{inc}}}_D-P^{t_{\textit{inc}}}_EV-\sum^{|\mathcal{I}|}_{i=1}(\underline{P_i}),\\
&\sum^{|\mathcal{I}|}_{i=1}(\overline{P_i})+P^{t_{\textit{dec}}}_EV-P^{t_{\textit{dec}}}_D-P^{t_{\textit{inc}}}_{SR}).
\end{aligned}
\end{equation}
The first term in the \textit{min} operator is the maximum increase range for $t_\textit{{inc}}$ and the second term is the maximum decrease range for $t_{\textit{dec}}$. With this range, we draw a random increase/decrease value $v\sim N(0, r/3)$. If the absolute value of $v$ is larger than $r$, this $v$ will be discarded and we randomly draw another one from the distribution. This process will iterate until a feasible $|v|\in[0, r]$ is drawn. This $v$ is then applied to modify the EV output values of the two previously selected time intervals, i.e.,
\begin{equation}\label{eqn:incdec}
\begin{aligned}
P^{t_{inc}}_{EV}\leftarrow P^{t_{inc}}_{EV}+|v|&\\
P^{t_{dec}}_{EV}\leftarrow P^{t_{dec}}_{EV}-|v|&.\\
\end{aligned}
\end{equation}
This operation may violate Constraints (\ref{eqn:capacity}) and (\ref{eqn:chargefreq}). In such cases, this modification on EV output values is reverted and the neighborhood search operator will do nothing in the current elementary reaction.

\subsection{Elementary Reactions}

In our proposed methodology, we employ the neighborhood search operator in all four elementary reactions, namely \textit{on-wall}, \textit{dec}, \textit{inter}, and \textit{syn}. For \textit{on-wall}, the neighborhood search operator can be employed as described before. For \textit{dec}, we first copy the input molecular structure to the two output molecules, and then perform neighborhood search on them separately. We treat \textit{inter} as two \textit{on-walls} occurring simultaneously. Finally, for \textit{syn}, we compare the performance of the two input solutions, pick the better one, and perform the neighborhood search on it.

\subsection{Economic Dispatch}

Up to now our solution is a binary matrix and a real-number vector. However the EVUC problem requires the power outputs of the units instead of the online status. So the algorithm must dispatch the load demand to all online units, and this process is called Economic Dispatch (ED) \cite{journals/tps/Chowdhury1990}. In EVUC, we use the lambda iteration method for economic dispatch in the UC problem as this method is guaranteed to find the optimal ED solution with a small enough estimation error \cite{journals/tps/Su2000}.

\section{Simulation Results and Discussion}

Our proposed approach was implemented in C++ on an Intel Core i5 3.1-GHz processor with MinGW compiler. We analyze the efficiency of V2G as well as the performance of CRO with a test system of up to 40 units.

\subsection{Testing Instance}

\begin{table}[t]
  \centering
  \caption{Capacity and Cost Coefficients of Thermal Units}
    \begin{tabular}{rrrrrr}
    \hline
    Unit  & $\overline{P_i}$(MW) & $\underline{P_i}$(MW) & $a_i$(\$/h) & $b_i$(\$/MWh) & $c_i$(\$/MWh$^2$) \\
    \hline
    1     & 455   & 150   & 1000  & 16.19 & 0.00048 \\
    2     & 455   & 150   & 970   & 17.26 & 0.00031 \\
    3     & 130   & 20    & 700   & 16.6  & 0.002 \\
    4     & 130   & 20    & 680   & 16.5  & 0.00211 \\
    5     & 162   & 25    & 450   & 19.7  & 0.00398 \\
    6     & 80    & 20    & 370   & 22.26 & 0.00712 \\
    7     & 85    & 25    & 480   & 27.74 & 0.0079 \\
    8     & 55    & 10    & 660   & 25.92 & 0.00413 \\
    9     & 55    & 10    & 665   & 27.27 & 0.00222 \\
    10    & 55    & 10    & 670   & 27.79 & 0.00173 \\
    \hline
    \end{tabular}
  \label{tab:unit1}
\end{table}

In our simulation, an independent system operator (ISO) of a 10-unit system is considered with 50 000 GVs. This ISO has been considered in many investigations \cite{journals/tie/AYSaber2011}\cite{journals/tsj/AYSaber2012}\cite{journals/tps/PHChen2012}. We consider a 24-hour scheduling horizon. Table \ref{tab:unit1} gives the capacity and cost coefficients of these thermal units and Table \ref{tab:unit2} gives the time-dependent parameters of these thermal units. In this system, the system reserve is set to 10\% of the total demand (load demand and EV charging demand), the shut down cost is ignored, and the start-up cost is calculated using
\begin{equation}\label{eqn:hotstart}
UC_i=
\left\{
\begin{aligned}
& UC^{\textit{hot}}_i & MDT_i\leq -\tau^t_i\leq MDT_i+T^{cold}_i\\
& UC^{\textit{cold}}_i & -\tau^t_i> MDT_i+T^{cold}_i
\end{aligned}
\right.
\end{equation}
where $T^{cold}_i$ is the extra time needed for unit $i$ to completely cool down besides $MDT_i$. So the start-up cost is temperature dependent where a cold unit requires $UC^{\textit{cold}}$ to start-up while a warm unit requires less cost $UC^{\textit{hot}}$. The load demands for the 24 hours are presented in Table \ref{tab:demand}. This load profile does not include the energy consumed by EVs.

\begin{table}[t]
  \centering
  \caption{Time-dependent Parameters of Thermal Units}
    \begin{tabular}{rrrrrrr}
    \hline
    Unit  & $MUT_i$ & $MDT_i$ & $\tau^1_i$(h) & $UC^{\textit{hot}}_i$(\$) & $UC^{\textit{cold}}_i$(\$) & $T^{\textit{cold}}_i$(h) \\
    \hline
    1     & 8     & 8     & 5     & 4500  & 9000  & 8 \\
    2     & 8     & 8     & 5     & 5000  & 10000 & 8 \\
    3     & 5     & 5     & 4     & 550   & 1100  & -5 \\
    4     & 5     & 5     & 4     & 560   & 1120  & -5 \\
    5     & 6     & 6     & 4     & 900   & 1800  & -6 \\
    6     & 3     & 3     & 2     & 170   & 340   & -3 \\
    7     & 3     & 3     & 2     & 260   & 520   & -3 \\
    8     & 1     & 1     & 0     & 30    & 60    & -1 \\
    9     & 1     & 1     & 0     & 30    & 60    & -1 \\
    10    & 1     & 1     & 0     & 30    & 60    & -1 \\
    \hline
    \end{tabular}
  \label{tab:unit2}
\end{table}

\begin{table}[t]
  \centering
  \caption{System Load Demand (Without EVs Demand, in MW)}
    \begin{tabular}{rrrrrrrrr}
    \hline
    Hour  & 1     & 2     & 3     & 4     & 5     & 6     & 7     & 8 \\
    Demand  & 700   & 750   & 850   & 950   & 1000  & 1100  & 1150  & 1200 \\
    \hline
    Hour  & 9     & 10    & 11    & 12    & 13    & 14    & 15    & 16 \\
    Demand & 1300  & 1400  & 1450  & 1500  & 1400  & 1300  & 1200  & 1050 \\
    \hline
    Hour  & 17    & 18    & 19    & 20    & 21    & 22    & 23    & 24 \\
    Demand & 1000  & 1100  & 1200  & 1400  & 1300  & 1100  & 900   & 800 \\
    \hline
    \end{tabular}
  \label{tab:demand}
\end{table}

In order to have a complete assessment of the proposed algorithm, we also made a 20- and 40-unit system by duplicating the 10-unit system and scaling the load demands as well as the system capacity (in terms of EV number) in proportion to the system size. For 20-unit system, there are 100 000 EVs and the load demand for the first hour is 1 400 MW. Such configuration is also studied in \cite{journals/tps/PHChen2012}.

The EV parameter values used in this paper are as follows: average EV battery capacity $\overline{EV^{\textit{cap}}}=15$kWh, charging frequency $\textit{freq}=1.0$, and average EV energy consumption over 24 hours $\overline{EV^{\textit{con}}}=8.22$kWh. All these numbers are adopted from \cite{journals/tie/AYSaber2011}. The parameter values for CRO to solve EVUC are listed in Table \ref{tab:param}. We select these parameter values using a trial-and-error method, which has been used in \cite{journals/tec/AYSLam2010}\cite{conference/cec/JJQYu2012PF}.

\begin{table}[t]
  \centering
  \caption{CRO Parameter Values}
    \begin{tabular}{rr}
    \hline
    Parameter & Value \\
    \hline
    Initial population size & 5 \\
    Initial molecular kinetic energy & 100 \\
    Initial central energy buffer size & 0 \\
    Collision rate & 0.05 \\
    Energy loss rate & 0.05 \\
    Decomposition threshold & 10 000 \\
    Synthesis threshold & 100 000 \\
    \hline
    \end{tabular}
  \label{tab:param}
\end{table}

\begin{table*}[t]
  \centering
  \caption{Best Scheduling and Dispatch of 10-Unit System without V2G using CRO}
    \begin{tabular}{r|rrrrrrrrrr|r|rr}
    \hline
    h\textbackslash{}unit & 1     & 2     & 3     & 4     & 5     & 6     & 7     & 8     & 9     & 10    & V2G   & Load & Reserve \\
    \hline
    1     & 455.00  & 324.98  & 0.00  & 0.00  & 0.00  & 0.00  & 0.00  & 0.00  & 0.00  & 0.00  & -79.98  & 700   & 16.67\% \\
    2     & 455.00  & 324.65  & 0.00  & 0.00  & 0.00  & 0.00  & 0.00  & 0.00  & 0.00  & 0.00  & -29.65  & 750   & 16.72\% \\
    3     & 455.00  & 324.33  & 130.00  & 0.00  & 0.00  & 0.00  & 0.00  & 0.00  & 0.00  & 0.00  & -59.33  & 850   & 14.37\% \\
    4     & 455.00  & 324.77  & 130.00  & 130.00  & 0.00  & 0.00  & 0.00  & 0.00  & 0.00  & 0.00  & -89.77  & 950   & 12.52\% \\
    5     & 455.00  & 324.19  & 130.00  & 130.00  & 25.00  & 0.00  & 0.00  & 0.00  & 0.00  & 0.00  & -64.19  & 1000  & 25.17\% \\
    6     & 455.00  & 360.00  & 130.00  & 130.00  & 25.00  & 0.00  & 0.00  & 0.00  & 0.00  & 0.00  & 0.00  & 1100  & 21.09\% \\
    7     & 455.00  & 410.02  & 130.00  & 130.00  & 25.00  & 0.00  & 0.00  & 0.00  & 0.00  & 0.00  & -0.02  & 1150  & 15.82\% \\
    8     & 455.00  & 455.00  & 130.00  & 130.00  & 30.01  & 0.00  & 0.00  & 0.00  & 0.00  & 0.00  & -0.01  & 1200  & 11.00\% \\
    9     & 455.00  & 455.00  & 130.00  & 130.00  & 85.00  & 20.00  & 25.00  & 0.00  & 0.00  & 0.00  & 0.00  & 1300  & 15.15\% \\
    10    & 455.00  & 455.00  & 130.00  & 130.00  & 162.00  & 33.00  & 25.00  & 10.00  & 0.00  & 0.00  & 0.00  & 1400  & 10.86\% \\
    11    & 455.00  & 455.00  & 130.00  & 130.00  & 162.00  & 73.00  & 25.00  & 10.00  & 0.00  & 10.00  & 0.00  & 1450  & 10.83\% \\
    12    & 455.00  & 455.00  & 130.00  & 130.00  & 162.00  & 80.00  & 25.00  & 43.00  & 10.00  & 10.00  & 0.00  & 1500  & 10.80\% \\
    13    & 455.00  & 455.00  & 130.00  & 130.00  & 162.00  & 33.01  & 25.00  & 10.00  & 0.00  & 0.00  & -0.01  & 1400  & 10.86\% \\
    14    & 455.00  & 455.00  & 130.00  & 130.00  & 85.02  & 20.00  & 25.00  & 0.00  & 0.00  & 0.00  & -0.02  & 1300  & 15.15\% \\
    15    & 455.00  & 455.00  & 130.00  & 130.00  & 30.00  & 0.00  & 0.00  & 0.00  & 0.00  & 0.00  & 0.00  & 1200  & 11.00\% \\
    16    & 455.00  & 323.64  & 130.00  & 130.00  & 25.00  & 0.00  & 0.00  & 0.00  & 0.00  & 0.00  & -13.64  & 1050  & 25.23\% \\
    17    & 455.00  & 324.47  & 130.00  & 130.00  & 25.00  & 0.00  & 0.00  & 0.00  & 0.00  & 0.00  & -64.47  & 1000  & 25.13\% \\
    18    & 455.00  & 360.06  & 130.00  & 130.00  & 25.00  & 0.00  & 0.00  & 0.00  & 0.00  & 0.00  & -0.06  & 1100  & 21.08\% \\
    19    & 455.00  & 440.01  & 130.00  & 130.00  & 25.00  & 20.00  & 0.00  & 0.00  & 0.00  & 0.00  & -0.01  & 1200  & 17.67\% \\
    20    & 455.00  & 455.00  & 130.00  & 130.00  & 162.00  & 33.01  & 25.00  & 0.00  & 0.00  & 10.00  & -0.01  & 1400  & 10.86\% \\
    21    & 455.00  & 455.00  & 130.00  & 130.00  & 85.02  & 20.00  & 25.00  & 0.00  & 0.00  & 0.00  & -0.02  & 1300  & 15.15\% \\
    22    & 455.00  & 455.00  & 130.00  & 0.00  & 35.00  & 0.00  & 25.00  & 0.00  & 0.00  & 0.00  & 0.00  & 1100  & 17.00\% \\
    23    & 455.00  & 324.79  & 130.00  & 0.00  & 0.00  & 0.00  & 0.00  & 0.00  & 0.00  & 0.00  & -9.79  & 900   & 14.31\% \\
    24    & 455.00  & 345.00  & 0.00  & 0.00  & 0.00  & 0.00  & 0.00  & 0.00  & 0.00  & 0.00  & 0.00  & 800   & 13.75\% \\
    \hline
    \multicolumn{14}{c}{Expected running cost = \$572467.30}\\
    \hline
    \end{tabular}
  \label{tab:bestlevelCRO}
\end{table*}

\begin{table*}[t]
  \centering
  \caption{Best Scheduling and Dispatch of 10-Unit System with V2G using CRO}
    \begin{tabular}{r|rrrrrrrrrr|r|rr}
    \hline
    h\textbackslash{}unit & 1     & 2     & 3     & 4     & 5     & 6     & 7     & 8     & 9     & 10    & V2G   & Load & Reserve \\
    \hline
    1     & 455.00  & 372.27  & 0.00  & 0.00  & 0.00  & 0.00  & 0.00  & 0.00  & 0.00  & 0.00  & -127.27  & 700   & 10.00\% \\
    2     & 455.00  & 372.26  & 0.00  & 0.00  & 0.00  & 0.00  & 0.00  & 0.00  & 0.00  & 0.00  & -77.26  & 750   & 10.00\% \\
    3     & 455.00  & 401.42  & 0.00  & 0.00  & 0.00  & 0.00  & 0.00  & 0.00  & 0.00  & 10.00  & -16.42  & 850   & 11.38\% \\
    4     & 455.00  & 455.00  & 0.00  & 0.00  & 39.99  & 0.00  & 0.00  & 0.00  & 0.00  & 0.00  & 0.01  & 950   & 12.84\% \\
    5     & 455.00  & 401.16  & 130.00  & 130.00  & 25.00  & 0.00  & 0.00  & 0.00  & 0.00  & 0.00  & -141.16  & 1000  & 16.72\% \\
    6     & 455.00  & 399.44  & 130.00  & 130.00  & 25.00  & 0.00  & 0.00  & 0.00  & 0.00  & 0.00  & -39.44  & 1100  & 16.90\% \\
    7     & 455.00  & 410.00  & 130.00  & 130.00  & 25.00  & 0.00  & 0.00  & 0.00  & 0.00  & 0.00  & 0.00  & 1150  & 15.83\% \\
    8     & 455.00  & 455.00  & 130.00  & 130.00  & 30.00  & 0.00  & 0.00  & 0.00  & 0.00  & 0.00  & 0.00  & 1200  & 11.00\% \\
    9     & 455.00  & 455.00  & 130.00  & 130.00  & 93.17  & 0.00  & 25.00  & 0.00  & 0.00  & 0.00  & 11.83  & 1300  & 10.00\% \\
    10    & 455.00  & 455.00  & 130.00  & 130.00  & 145.89  & 20.00  & 25.00  & 0.00  & 0.00  & 0.00  & 39.11  & 1400  & 10.00\% \\
    11    & 455.00  & 455.00  & 130.00  & 130.00  & 145.90  & 20.00  & 25.00  & 0.00  & 0.00  & 0.00  & 89.10  & 1450  & 10.00\% \\
    12    & 455.00  & 455.00  & 130.00  & 130.00  & 154.25  & 20.00  & 25.00  & 10.00  & 0.00  & 0.00  & 120.75  & 1500  & 12.53\% \\
    13    & 455.00  & 455.00  & 130.00  & 130.00  & 145.89  & 20.00  & 25.00  & 0.00  & 0.00  & 0.00  & 39.11  & 1400  & 10.00\% \\
    14    & 455.00  & 455.00  & 130.00  & 130.00  & 84.99  & 20.00  & 25.00  & 0.00  & 0.00  & 0.00  & 0.01  & 1300  & 15.15\% \\
    15    & 455.00  & 455.00  & 130.00  & 130.00  & 30.00  & 0.00  & 0.00  & 0.00  & 0.00  & 0.00  & 0.00  & 1200  & 11.00\% \\
    16    & 455.00  & 400.90  & 130.00  & 130.00  & 25.00  & 0.00  & 0.00  & 0.00  & 0.00  & 0.00  & -90.90  & 1050  & 16.75\% \\
    17    & 455.00  & 401.24  & 130.00  & 130.00  & 25.00  & 0.00  & 0.00  & 0.00  & 0.00  & 0.00  & -141.24  & 1000  & 16.72\% \\
    18    & 455.00  & 403.51  & 130.00  & 130.00  & 25.00  & 0.00  & 0.00  & 0.00  & 0.00  & 0.00  & -43.51  & 1100  & 16.48\% \\
    19    & 455.00  & 440.07  & 130.00  & 130.00  & 25.00  & 20.00  & 0.00  & 0.00  & 0.00  & 0.00  & -0.07  & 1200  & 17.66\% \\
    20    & 455.00  & 455.00  & 130.00  & 130.00  & 145.91  & 20.00  & 25.00  & 0.00  & 0.00  & 0.00  & 39.09  & 1400  & 10.00\% \\
    21    & 455.00  & 455.00  & 130.00  & 130.00  & 85.00  & 20.00  & 25.00  & 0.00  & 0.00  & 0.00  & 0.00  & 1300  & 15.15\% \\
    22    & 455.00  & 455.00  & 130.00  & 0.00  & 35.00  & 0.00  & 25.00  & 0.00  & 0.00  & 0.00  & 0.00  & 1100  & 17.00\% \\
    23    & 455.00  & 360.45  & 130.00  & 0.00  & 0.00  & 0.00  & 0.00  & 0.00  & 0.00  & 0.00  & -45.45  & 900   & 10.00\% \\
    24    & 455.00  & 372.27  & 0.00  & 0.00  & 0.00  & 0.00  & 0.00  & 0.00  & 0.00  & 0.00  & -27.27  & 800   & 10.00\% \\
    \hline
    \multicolumn{14}{c}{Expected running cost = \$564727.87}\\
    \hline
    \end{tabular}
  \label{tab:bestsmartCRO}
\end{table*}

\subsection{Analysis on V2G Efficiency}

In order to demonstrate the advantage of our V2G scheme compared with the current EV charging-only scheduling scheme, we propose and investigate two models representing the two schemes, respectively.

\subsubsection{Load-Leveling Model}

In this model, EVs are charged through thermal units using load-leveling optimization. No V2G operations are made. In this model, Constraint (\ref{eqn:chargefreq}) in EVUC becomes
\begin{equation}\label{eqn:loadlevel}
P^t_{EV}\leq0\hspace{1em}t=1,2,\cdots,T.
\end{equation}
The simulation results for this model is presented in Table \ref{tab:bestlevelCRO}.

\subsubsection{V2G Model}

In this model, EVs are charged through thermal units as loads and discharged to the grid as sources. The simulation results for this model is presented in Table \ref{tab:bestsmartCRO}.

From Tables \ref{tab:bestlevelCRO} and \ref{tab:bestsmartCRO} we can see that, the total running cost is reduced by \$7738.63 every 24-hour cycle, due to V2G operations. This phenomenon can also be observed in the simulations on the 20- and 40-unit systems, whose results are presented in Table \ref{tab:compmodel}. All the simulation results show that introducing V2G technology to existing power system can effectively reduce the running cost.

\begin{table}[t]
  \centering
  \caption{Best Running Cost Comparison between Two Models}
    \begin{tabular}{rrrr}
    \hline
    Units & Load Leveling & V2G & difference \\
    \hline
    10    & \$572,467.30  & \$564,727.87  & -\$7,739.43 \\
    20    & \$1,145,196.73  & \$1,128,131.28  & -\$17,065.45 \\
    40    & \$2,286,394.59  & \$2,257,690.96  & -\$28703.63 \\
    \hline
    \end{tabular}
  \label{tab:compmodel}
\end{table}

\subsection{Comparing CRO with Other Metaheuristics}

In order to demonstrate the superiority of CRO in solving EVUC, we compare the simulation result of CRO with other metaheuristics on 10-, 20-, and 40-unit systems. The selected metaheuristics are all algorithms with excellent performance in solving UC and related problems. These are EP \cite{journals/tps/Juste1999}, QIEA \cite{journals/tps/Lau2009}, SA \cite{conference/psce/Simo2006}, LRPSO \cite{conference/pmaps/Logen2010}, and ES-EPSO \cite{journals/tps/PHChen2012}. As there is no published results on our proposed EVUC problem, we implement these algorithms according to the description in the corresponding literature. The function evaluation limit is set to 50 000. The parameter values are selected according to the published records. Every algorithm is tested over all systems for 100 times. The simulation results are presented in Table \ref{tab:simres}.

\begin{table}[t]
  \centering
  \caption{Comparison of Solution Performance of CRO and Other Algorithms}
    \begin{tabular}{rrrrr}
    \hline
    Units & Algorithm & Best Cost(\$) & Mean Cost(\$) & Mean Time(s) \\
    \hline
    10    & CRO   & \textbf{564,727.87} & \textbf{565,019.42} & 2.02 \\
          & EP    & 566,016.44  & 569,217.98  & 1.99 \\
          & QIEA  & 565,294.13  & 565,364.46  & 2.17 \\
          & SA    & 567,639.85  & 568,249.21  & 2.09 \\
          & LRPSO & 566,912.80  & 567,438.57  & 2.85 \\
          & ES-EPSO & 565,047.61  & 565,497.39  & 2.96 \\
    \hline
    20    & CRO   & \textbf{1,128,131.28} & \textbf{1,129,473.01} & 3.48 \\
          & EP    & 1,131,524.73  & 1,136,132.33  & 3.44 \\
          & QIEA  & 1,130,148.48  & 1,130,578.16  & 3.7 \\
          & SA    & 1,134,861.47  & 1,136,905.79  & 3.61 \\
          & LRPSO & 1,133,126.98  & 1,133,913.37  & 5.41 \\
          & ES-EPSO & 1,129,632.35  & 1,130,975.40  & 5.69 \\
    \hline
    40    & CRO   & \textbf{2,257,690.96} & \textbf{2,259,279.49} & 5.99 \\
          & EP    & 2,263,546.88  & 2,272,957.50  & 5.95 \\
          & QIEA  & 2,260,964.88  & 2,261,157.61  & 6.31 \\
          & SA    & 2,269,970.59  & 2,273,957.16  & 6.22 \\
          & LRPSO & 2,266,485.37  & 2,267,800.28  & 10.32 \\
          & ES-EPSO & 2,259,141.88  & 2,261,421.76  & 10.8 \\
    \hline
    \end{tabular}
  \label{tab:simres}
\end{table}

From the results we can see CRO outperforms other algorithms in every test on both the comparison of best cost and the mean cost. The superiority is enhanced when the problem size increases. In terms of computational time, EP is the fastest algorithm but the advantage over CRO is negligible. Almost 95\% the of total time of CRO, EP, QIEA, and SA is employed to solve ED using the lambda iteration method, which is not avoidable in all UC simulations. As to LRPSO and ES-EPSO, the relatively high computational complexities of the algorithms make them less competitive.

\section{Conclusion}

In this paper, we propose a new optimization problem, namely, joint scheduling of EVs and UC, called EVUC. Our formulation can overcome the drawbacks of previous formulations. The main idea of the problem is to employ EVs as power sources and storages at different times, instead of only using them as loads. The major improvement of our formulation with previous formulations is that we consider the special characteristics of EVs while optimizing the total system running cost. This improvement makes our model more realistic and also more effective at reducing the total system running cost. In order to assess the efficiency of our formulation, we employ CRO to solve the optimization problem. The simulation results indicate that our proposed scheduling algorithm can significantly reduce the running cost while maintaining sufficient spinning reserve to handle emergency situations. Moreover, we compare the simulation results of CRO with a wide range of other metaheuristics with excellent performance in solving similar problems in previous literature. CRO outperforms all other compared metaheuristics in terms of both the best cost and the mean cost, and the simulation time needed is among the shortest. All these phenomenon show that CRO is an efficient method for our proposed EVUC problem.

\bibliographystyle{IEEEtran}
\bibliography{EV_UC_CRO}
\end{document}